\pdfoutput=1

\documentclass[11pt]{article}

\usepackage[]{EMNLP2023}
\usepackage{times}
\usepackage{latexsym}
\usepackage{amsmath}
\usepackage[T1]{fontenc}

\usepackage[utf8]{inputenc}

\usepackage{microtype}

\usepackage{inconsolata}
\usepackage{graphicx}

\title{AdapterDistillation: Non-Destructive Task Composition with Knowledge Distillation}

\author{Junjie Wang\textsuperscript{*}, Yicheng Chen\textsuperscript{*}, Wangshu Zhang, Sen Hu, Teng Xu, Jing Zheng \\
        Ant Group \\ \{benge.wjj, yicheng.chen, wangshu.zws, hs272483, harvey.xt, jing.zheng\}@antgroup.com}

\begin{document}
\maketitle
\let\thefootnote\relax\footnote{*Equal Contributions}
\begin{abstract}
Leveraging knowledge from multiple tasks through introducing a small number of task specific parameters into each transformer layer, also known as adapters, receives much attention recently. However, adding an extra fusion layer to implement knowledge composition not only increases the inference time but also is non-scalable for some applications. To avoid these issues, we propose a two-stage knowledge distillation algorithm called AdapterDistillation. In the first stage, we extract task specific knowledge by using local data to train a student adapter. In the second stage, we distill the knowledge from the existing teacher adapters into the student adapter to help its inference. Extensive experiments on frequently 
asked question retrieval in task-oriented dialog systems validate the efficiency of AdapterDistillation. We show that AdapterDistillation 
outperforms existing algorithms in terms of accuracy, resource consumption and inference time.

\end{abstract}

\section{Introduction}

Recently task-oriented dialogue systems have found extensive applications in diverse business domains \citep{yan2017building, DBLP:conf/acl/WeiLPTCHWD18, DBLP:conf/acl/ValizadehP22}. Owing to the idiosyncratic features of these domains, custom dialogue systems are often required. Nonetheless, the fundamental functions and architectures underlying these systems typically exhibit noteworthy similarities. Hence, the adoption of a platform-based strategy for accommodating task-oriented dialogue systems across multiple domains has emerged as a promising and effective approach. 

\begin{figure}[tb!]
	\centering
	\includegraphics[width=0.48\textwidth]{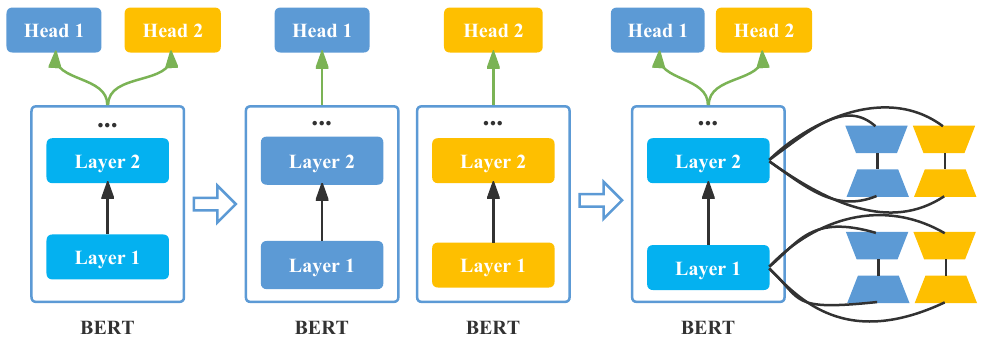}
	\caption{Three popular multi-tenant learning methods. Left: Multi-Task Learning. Middle: Independent Learning. Right: Adapter Learning.}
	\label{fig:intro}
	\vspace{-10pt}
\end{figure}
One popular method is called Multi-Task Learning (MTL), which aims to train multiple tasks simultaneously based on the shared representation of all tasks as shown in Figure \ref{fig:intro}, resulting in relatively good performance on each task \citep{collobert2008unified,DBLP:journals/jstsp/ChenBTS22,DBLP:conf/ciss/ChenBS22}. However, new tenants often register on the platform in a streaming manner. Therefore, predictions for the existing tenants in MTL would be compromised when new tenants are added to the platform since retraining often occurs at that time. 


To ensure that tenants do not interfere with each other, an intuitive approach is to train a task-specific model for each tenant. However, this independent training approach requires a significant amount of resources to store complete model parameters. It is clear that the resource consumption becomes the bottleneck as many tenants register on the platform. Additionally, fine-tuning all parameters on a tenant with very little data can often lead to severe overfitting \citep{DBLP:journals/csur/Dietterich95, DBLP:journals/jcisd/Hawkins04}. Thus, providing tenants with the ability to solve designated tasks
with limited resources is necessary.


Owing to the distinctive properties of platform-based systems, tasks implemented on the platform had better satisfy the following criteria: 1) The platform witnesses a continuous influx of new tenants. It is incumbent upon the model to ensure the performance of the existing tenants are not destructed when new tenants are added.  2) The resources of the platform are limited, thereby necessitating the provision of tenants with the capability to ensure the task performance of each tenant with minimal storage and computational resources. Given this practical limitation, for an incoming tenant, how to utilize the current tenant data and the existing tenant models (also called teacher models) becomes an interesting topic.

In order to maintain the low-resource and scalability of the model while utilizing the existing tenant knowledge, we propose an algorithm called AdapterDistillation. In AdapterDistillation, we employ
adapters to capture task specific features by adding a few extra parameters in the transformer layer, and then distillate knowledge from the existing teacher adapters into the current student adapter. To be exact, when a new tenant comes, we first train an adapter module based on its own local data. Then we load the adapter modules of all the current teacher adapters to assist the training of this new student tenant through knowledge distillation.

\textbf{Our contributions:}
The proposed approach has several advantages:  1) Fusion is only added during the second stage of training to guide the current student adapter learning and is not required during inference, ensuring structure consistency between the student adapter and the existing teacher adapters. 2) Since the adapter structure is consistent during prediction and no additional parameters are required, the scalability and low-resource nature of the model itself are retained. To summarize, our contributions are:

\begin{itemize}
\item We formulate the construction of a platform-based multi-task problem as a transfer learning problem and leverage the low-resourced adapter model to handle this.
\item We  propose an AdapterDistillation algorithm which guarantees low-resources and scalability while utilizing the existing tenant knowledge to improve performance.
\item We verify noteworthy enhancements of the proposed AdapterDistillation algorithm in terms of accuracy, resource consumption and inference time, using a Frequently
Asked Question (FAQ) retrieval service in task-oriented
dialog systems.
\end{itemize}

\section{Relevant Work}

Recently, adapters have been proposed to capture task-specific features while maintaining similar results to fine-tuning all parameters, which has been widely applied to downstream tasks such as machine translation and cross-lingual transfer \citep{houlsby2019parameter, 
DBLP:conf/emnlp/PfeifferVGR20,
DBLP:conf/acl/LiL20, DBLP:conf/emnlp/LesterAC21, DBLP:conf/iclr/HeZMBN22}. Specifically, adapters insert two bottleneck modules into each transformer layer of the pre-trained model \citep{houlsby2019parameter}. During training, all parameters of the pre-trained model are frozen, and only the parameters of the newly added modules are trained. Some researchers \citep{DBLP:conf/emnlp/PfeifferRPKVRCG20,DBLP:conf/emnlp/PfeifferVGR20} has further improved the insertion position of adapters through structural search, reducing the number of adapter insertions and thus minimizing the increase in parameter quantity and inference speed. A new type of adapters called Lora \citep{DBLP:conf/iclr/HuSWALWWC22} has been proposed to first perform low-rank decomposition on the model parameters and then insert adapters into the key, query, and value matrices of each attention layer. This approach enhances performance and enables parallel execution of the adapter module, thus reducing inference time.
Due to the lack of clarity regarding the inter-dependencies and key success factors of various adapter methods, He et al. \citep{DBLP:conf/iclr/HeZMBN22} dissected the design of several classic adapter algorithms and established a unified framework to explore the connections between different adapter methods. Note that all of the work discussed in this paragraph is complimentary to the proposed method called AdapterDistillation since our algorithm is not limited to a certain type of adapters.  Thus we can combine our developed approach with all of the work discussed in this paragraph 
to obtain extra gains. 

In addition to optimizing the structure and position of adapters for each individual task, adding extra components on the top of multiple adapters to maximize the transfer of knowledge across multiple tasks is another efficient way to enhance the performance on each task. For example, via adding an extra fusion layer, the AdapterFusion method is proposed to effectively share knowledge across multiple tasks while balancing the various target tasks \citep{DBLP:conf/eacl/PfeifferKRCG21}. To be specific, this method uses a two-stage training approach: first, train the corresponding adapter for each task, then load all adapters simultaneously and freeze them, and train an additional adapter fusion layer to aggregate the outputs of all adapters, allowing the model to implicitly and automatically learn to utilize knowledge from different tasks. But this approach faces some challenges in practical applications: since the parameter size of the fusion layer is linearly related to the number of loaded adapters, when the number of adapters is too large, a lot of resources will be used for fusion such that the purpose of using adapters gets lost. Additionally, adding a fusion layer after the adapters leads to larger inference time, resulting in a worse user experience.

To efficiently utilize existing task knowledge and meet the requirement of the platform for streaming task integration, we propose a plug-and-play AdapterDistillation algorithm. By fusing and distilling the knowledge of existing tasks into the current task during training, we can keep the model structure and inference speed unchanged while achieving comparable results to AdapterFusion.

\section{Problem Definition}
As we know, training adapters for $N$ tasks in parallel might not be practical since tenants often register on the platform in a streaming manner. Based on this fact, the time for the $j$-th task registered on the platform is assumed to be earlier than that for the $i$-th task, that is $t_j<t_i$, when $j<i$ for an ordered collection of N tasks denoted as $T=\{t_1,t_2,...,t_N \}$. 

Throughout the paper, we have the following settings which are typically true in practice:
\begin{enumerate}
    \item The task considered in this paper is non-destuctive, which means when the  $N$-th task is registered on the platform, the performance of the previous $(N-1)$ tasks $\{t_1,t_2,...,t_{N-1} \}$ should not be impacted.
    \item Since the platform often has limited resources, it is reasonable to assume every task needs to be solved with limited computing and memory resources.
    \item Due to privacy and security issues,  corpus of labeled text for  the  $N$-th task is only available locally.
\end{enumerate}
Based on the above setting, in this paper we are aimed at maximizing the transfer of knowledge from the existing tasks to the current new task without impacting the existing tasks, which is more suitable for a practical scenario where each task is registered on the platform in a streaming manner.  


\section{AdapterDistillation}
AdapterFusion allows sharing of information between different tasks through an extra fusion layer at the cost of longer inference time and larger fusion layer \citep{DBLP:conf/eacl/PfeifferKRCG21}. However, as a new task is registered on the platform, the existing tasks will be impacted in. In order to mitigate this, 
we propose AdapterDistillation to allow sharing of information from the existing tasks to the new one while avoiding the impact of the existing tasks without increasing inference time.

\subsection{Adapter Learning and Distillation Algorithm}
The proposed AdapterDistillation algorithm is a two-stage learning algorithm. In the first stage, we train an adapter model $\phi_N^{first}$ for the $N$-th new task when it is registered on the platform based on its own local data. 

In the second stage, we employ knowledge distillation to transfer knowledge from the existing tasks to the new adapter, which means the parameter weight of this new adapter will be updated in the second stage. To be exact, assuming there have been $(N-1)$ adapters registered on the platform with their weights being denoted as $\{\phi_n\}_{n=1}^{N-1}$ and the $N$-th adapter with its weight trained in the first stage being denoted as $\phi_N^{first}$. With a fixed pretrained Bert-based model $\Theta$ and the existing adapters $\{\phi_n\}_{n=1}^{N-1}$ and $\phi_N^{first}$, the data fusion related parameters $\Omega$ and the $N$-th adapter weight $\phi$ have been introduced to learn how to distill knowledge from the existing adapters $\{\phi_n\}_{n=1}^{N-1}$ and $\phi_N^{first}$ to better solve the $N$-th task. The training process can be represented as
\begin{align}\label{distill_explain}
&\Omega_N, \phi_N \gets \mathop{\arg\min}\limits_{\Omega,\phi} CrossEntropy(D_N;\phi,\Theta) 
  \notag\\
&\! +\!\eta\cdot \! Distill(D_N;
\{\phi_n\}_{n=1}^{N-1},\phi_N^{first},\Omega,\phi,\Theta)
\end{align}
where $D_N$ are corpus of labeled text for task $N$, $\eta$ is a predefined constant to balance the distillation loss and  the binary cross entropy loss, $\Omega_N$ is a set of newly learned fusion-related parameters to transfer the existing knowledge from the existing adapters to the $N$-th adapter for task $N$, and $\phi_N$ is the final weight for adapter $N$. It is worth mentioning that similar to AdapterFusion, each adapter in AdapterDistillation will be trained twice where the second stage is mainly aimed at implementing knowledge composition. However, different from AdapterFusion which keeps the fusion layer during the inference time, AdapterDistillation will only employ the $N$-th adapter module to do inference without adding an extra fusion layer (shown in Figure. \ref{fig:adapterdistill_architecture}) which leads to faster inference time without impacting the performance of the existing tasks. This makes sense since the $N$-th adapter weight $\phi_N$ already contains the sharing of information between $N$ tasks after knowledge distillation.

\subsection{Detailed Components}

During the training process, AdapterDistillation learns to distill the knowledge from the existing $(N-1)$ adapters to the $N$-th adapter by introducing the fusion weights $\Omega$ and updating the $N$-th adapter weights $\phi$. The fusion weights $\Omega$ transfer the existing knowledge to the $N$-th adapter module by dynamically introducing a distillation loss term as shown in (\ref{distill_explain}). This will push the $N$-th adapter to learn knowledge not only from its own task data $D_N$ but also from the previous $(N-1)$ adapter intermediate representations.

\begin{figure}[tb!]
	\centering	\includegraphics[width=0.49\textwidth]{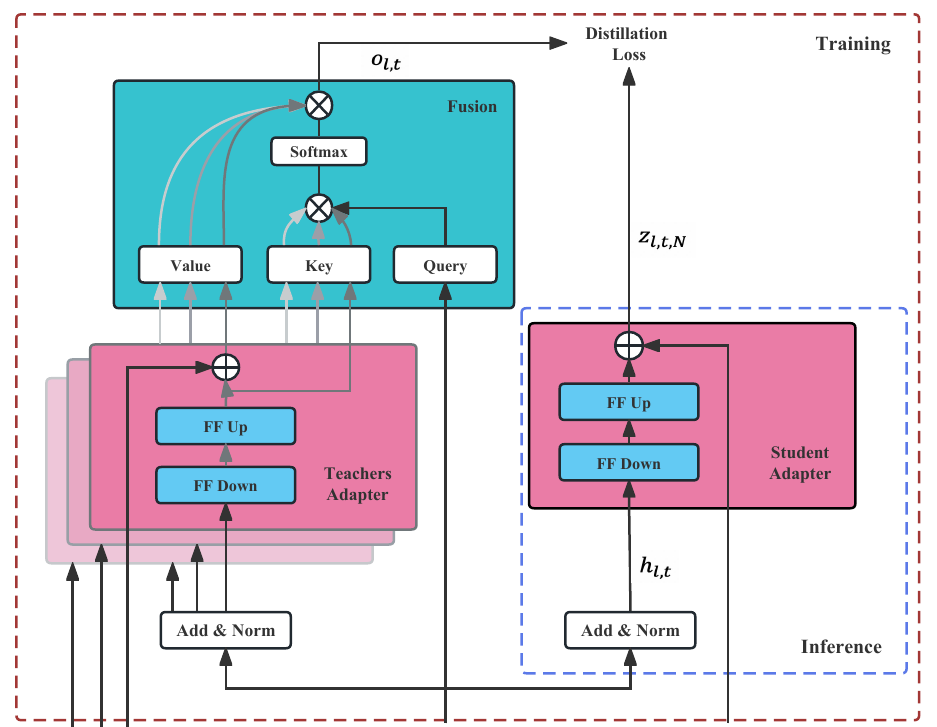}
	\caption{ Our AdapterDistillation architecture. This includes trainable weights Query ${\bf{Q}}_l$, Key ${\bf{K}}_l$, Value ${\bf{V}}_l$ and the $N$-th adapter weight $\phi_N$ at each transformer layer $l$.}
	\label{fig:adapterdistill_architecture}
	\vspace{-10pt}
\end{figure}

As shown in Figure \ref{fig:adapterdistill_architecture}, our AdapterDistillation architecture contains three components, that is, an adapter fusion, $N$ teacher adapters and a $N$-th student adapter. In the student adapter part, the output of the feed-forward sublayer of layer $l$ at iteration $t$, denoted as $\mathbf{h}_{l, t}$, is fed into the $N$-th adapter to obtain the $N$-th adapter output $ \mathbf{z}_{l,t,N}=g(\mathbf{h}_{l,t},\phi)$ with $g(\mathbf{h}_{l,t},\phi)$ being the nonlinear transformation and $\phi$ being the adapter parameters to be optimized. Interestingly enough, in the $N$ teacher adapters, we not only use the previous $(N-1)$ fully trained adapters $\phi_n$ as teacher adapters to enable sharing of information between different tasks but also add the $N$-th partially trained adapter $\phi_N^{first}$ obtained in the first stage as a teacher adapter to insert some task specific knowledge. In other words, the output of $N$ teacher adapters can be denoted as $ \mathbf{z}_{l,t,n}=g(\mathbf{h}_{l,t},\phi_n)$ for $n=1,2,…,N-1$ and $ \mathbf{z}_{l,t,N}=g(\mathbf{h}_{l,t},\phi_N^{first})$. 

Similar to AdapterFusion \citep{DBLP:conf/eacl/PfeifferKRCG21}, our AdapterDistillation dynamically combines different adapters by introducing Query ${\bf{Q}}_l$, Key ${\bf{K}}_l$, and Value ${\bf{V}}_l$ at each transformer layer $l$ with its complete set being $\Omega=\{{\bf{Q}}_l,{\bf{K}}_l,{\bf{V}}_l\}_{l=1}^L$. We employ $\mathbf{z}_{l,t,n}$ for $n=1,2,...,N$ as the input to the value and key transformation to obtain $\mathbf{z}_{l, t, n}^{v} =\mathbf{z}_{l, t, n}^{\top} \mathbf{V}_l$ and $\mathbf{z}_{l, t, n}^{k} =\mathbf{z}_{l, t, n}^{\top} \mathbf{K}_l$, respectively. The output of the feed-forward sublayer $\mathbf{h}_{l,t}$ is used as input to the query transformation to obtain $\mathbf{h}_{l, t}^Q=\mathbf{h}_{l, t}^{\top} \mathbf{Q}_l$. Then the output of the adapter fusion $\mathbf{o}_{l, t}$ can be obtained as 
\begin{align}
\mathbf{o}_{l, t} = p_{l,t}^T\mathbf{Z}_{l, t, n}^{v}
\end{align}
with $p_{l,t}=\mathrm{softmax}(\mathbf{h}_{l, t}^Q \bigotimes \mathbf{z}_{l, t, n}^{k})$ and $\mathbf{Z}_{l, t, n}^{v} = [\mathbf{z}_{l, t, 1}^{v},\mathbf{z}_{l, t, 2}^{v},...,\mathbf{z}_{l, t, N}^{v}]$. Note that we employ $p_{l,t}$ to learn to weight the adapters with regard to the context.
Finally, the distillation loss described in (\ref{distill_explain}) is defined as 
\begin{align}
Distill = \|\mathbf{o}_{l, t}-\mathbf{z}_{l, t, N}\|.
\end{align}
It is worth mentioning that in the second stage we jointly optimize the adapter fusion $\Omega$ and $\phi$ so as to obtain the optimal $\phi_N$ which contains the most useful mixed knowledge from available adapters. Then during the inference stage, we only employ $\phi_N$ to implement the prediction for the $N$-th task without considering the adapter fusion part so as to reduce inference time. On the other hand, only employing $\phi_N$ can also lead to comparable performance as AdapterFusion, which will be shown next.

\section{Experiments}
\begin{table*}[htbp]
\setlength{\tabcolsep}{4pt}
\renewcommand{\arraystretch}{1.3}
\begin{tabular}{lccccc}
\hline
Dataset &
  \begin{tabular}[c]{@{}c@{}}Full\end{tabular} &
  \begin{tabular}[c]{@{}c@{}}HEAD\end{tabular} &
  \begin{tabular}[c]{@{}c@{}} Adapter\end{tabular} &
  \begin{tabular}[c]{@{}c@{}} AdapterFusion\end{tabular} &
  \begin{tabular}[c]{@{}c@{}} AdapterDistill\end{tabular} \\ \hline
International Payments              & 76.4(0.859) & 64.2(0.714) & 74.8(0.834)          & 76.2(0.855) & 75.3(0.84) \\
Merchant Payments              & 88.74(0.953) & 84.03(0.911) & 84.71(0.924)          & 85.21(0.936) & 85.21(0.928) \\
Broadband Installation              & 100(1.0) & 94.74(0.995) & 96.49(0.997)          & 98.25(1.0) & 96.49(0.997) \\
Cross-border Payments              & 82.4(0.885) & 68.7(0.763) & 77.3(0.849)          & 79.9(0.875) & 77.6(0.851) \\
Merchant Signing              & 80.39(0.894) & 80.39(0.887) & 82.35(0.894) & 76.47(0.902) & 84.31(0.907) \\
Human Resources              & 87.02(0.919) & 68.11(0.685) & 75.17(0.808)          & 81.32(0.869) & 79.73(0.825) \\
IT Support              & 99.28(0.999) & 76.2(0.855) & 93.51(0.982)          & 96.88(0.993) & 94.95(0.988) \\
Administration              & 95.77(0.984) & 71.96(0.817) & 93.65(0.975)          & 92.59(0.976) & 94.71(0.976) \\ \hline
Average                & 88.75(0.937) & 76.04(0.828) & 84.75(0.908)          & 85.85(0.926) & \textbf{86.04(0.914)} \\ \hline
Added Params Per Task      & 100\% & 0.01\% & 1.45\%          & 21.36\% & \textbf{1.45\%} \\ \hline
\end{tabular}
\caption{The accuracy and AUC for the 10-th tenant with different architectural setups. The result within the parenthesis is AUC. Added Params Per Task represents the percentage of additional parameters added for each task.}
\label{acc}
\end{table*}

To validate the effectiveness of AdapterDistillation in terms of accuracy, resource consumption and inference time, its performance is evaluated through a practical scenario where Frequently Asked Question (FAQ) retrieval is considered in task-oriented dialog systems. 

\subsection{Experimental Setting}
To benchmark AdapterDistillation, we compare with the following four model architectures, namely, BERT + adapters (abbr. as Adapter), fullly fine-tuning BERT model (abbr. as Full), head-only fine-tuning BERT model (abbr. as HEAD) and AdapterFusion. A detailed experimental setting can be found in Appendix \ref{Experimental_Setting}. 


\subsection{Datasets and Metrics}
We select 9 existing tenant models from the platform as teacher adapters, covering fields such as medical care, transportation, insurance, shopping, photography, lease and et al, and employ the performance of the 10-th tenant (student adapter) to evaluate the considered models. In order to reduce the variance, we independently choose 8 unregistered tenants from different business domains as the 10-th student tenant. The 8 independent student tenants are from international payments, merchant payments, broadband installation, cross-border payments, merchant signing, human resources, administrative management, and IT support. 
The data size for each student tenant ranges from 1000 to 5000 which has been  divided into the training, validation, and test dataset with the ratio being 8:1:1. It is worth mentioning that we not only use accuracy and AUC to evaluate the performance, but also use the resource consumption and inference time as additional metrics to indicate the functionality of the models of interest for online practical applications.

\subsection{Performance}

As shown in Table \ref{acc}, it is straightforward to see that Full fine-tuning leads to much better performance compared to training only the HEAD (12.71\% accuracy increase) at the cost of adding more trainable parameters. Additionally, adapters achieve a little worse accuracy performance compared to Full fine-tuning but with only the 1.45\% extra added parameters, which is promising. Table \ref{acc} also shows that AdapterFusion can reduce the performance gap by adding an extra fusion layer to implement knowledge composition but at the cost of adding 21.36\% extra parameters. Interestingly, the overall accuracy of the propose AdapterDistillation method achieves even better accuracy than AdapterFusion but with only much fewer added parameters (21.36\% VS 1.45\%). This makes sense since the representations from several such teacher adapters have been inserted into the student adapter through knowledge distillation, which means the fusion layer is not that important for AdapterDistillation during the inference stage.

\begin{figure*}[tb!]
	\centering
	\includegraphics[width=1.0\textwidth]{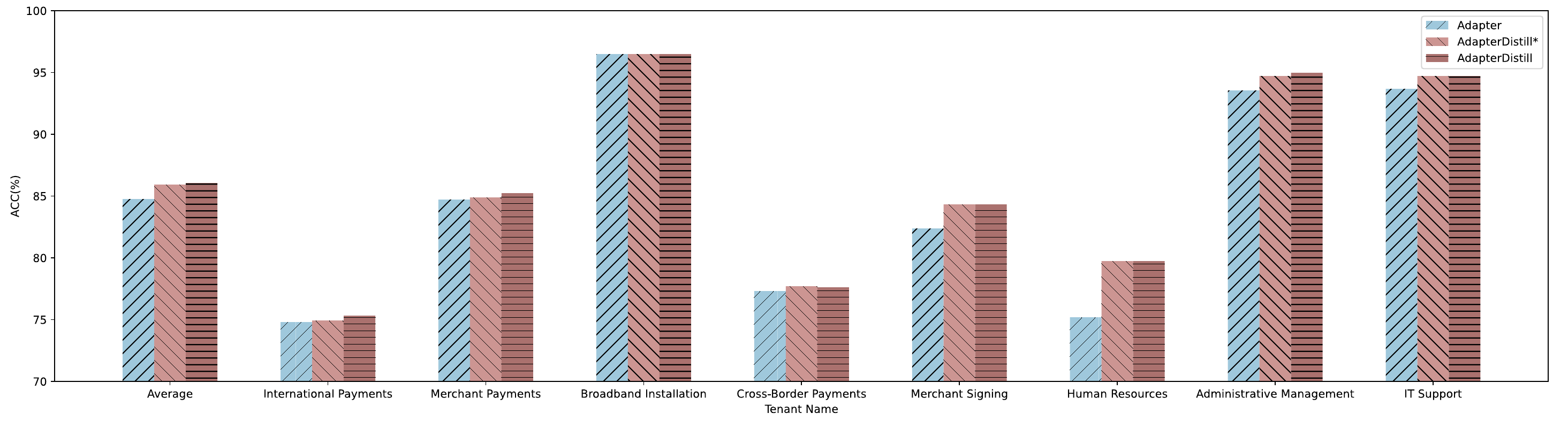}
	\caption{Accuracy of the 10-th tenant from 8 different domains for the different architectural setups. AdapterDistill* is the algorithm where we remove the current tenant as a teacher adapter in the second stage. }
	\label{fig:ablation_1}
	\vspace{-10pt}
\end{figure*}

\begin{table}[htb]
\setlength{\tabcolsep}{3.5pt}
\renewcommand{\arraystretch}{1.1}
\centering
\begin{tabular}{ccccc}
\hline
  \begin{tabular}[c]{@{}c@{}}Storage\\ Space\end{tabular} &
  \begin{tabular}[c]{@{}c@{}}BERT\\ Fine-Tune\end{tabular} &
  \begin{tabular}[c]{@{}c@{}}Adapter\\ Fusion\end{tabular} &
  \begin{tabular}[c]{@{}c@{}}Adapter\\ Distill\end{tabular} \\ \hline
500MB & 1   & 0       & \textbf{18}    \\
1GB   & 2   & 6     & \textbf{109}   \\
5GB   & 13  & 53   & \textbf{815}   \\
10GB  & 26  & 111    & \textbf{1698}  \\
50GB  & 130 & 578   & \textbf{8760}  \\
100GB & 261 & 1161  & \textbf{17587} \\ \hline
\end{tabular}
\caption{The number of tenants that can be supported by different methods versus the storage space. 
}

\label{tenant_num}
\end{table}


In addition to accuracy and AUC, resource consumption is an important indicator of deployment costs.
In terms of storage space required during the inference stage, the pre-trained Bert-base-Chinese model takes up approximately 391MB. The fusion module and the adapter module occupies 82MB and 3.5MB, respectively. The last classification layer requires 2.3MB. As a result, in addition to the base model, it takes approximately extra 119.3MB for AdapterFusion but only takes approximately extra 5.8MB for AdapterDistillation when the 10-th tenant registers on the platform. In Table \ref{tenant_num}, we consider the number of tenants that can be supported by different methods. It shows that when the storage space is 100GB, AdapterDistillation can support 67 times more tenants
than Full fine-tuning and 20 times more tenants than
AdapterFusion. The results in Table \ref{tenant_num} indicate that AdapterDistillation has significant advantages on resource consumption compared to others, which becomes more pronounced as the storage space becomes larger.


\begin{table}[]
\setlength{\tabcolsep}{1.75pt}
\renewcommand{\arraystretch}{1.1}
\begin{tabular}{cccc}
\hline
\begin{tabular}[c]{@{}c@{}}Batch \\ Size\end{tabular} &
  \begin{tabular}[c]{@{}c@{}}BERT \\ Fine-Tune\end{tabular} &
  \begin{tabular}[c]{@{}c@{}}Adapter\\ Fusion\end{tabular} &
  \begin{tabular}[c]{@{}c@{}}Adapter/\\ AdapterDistill\end{tabular} \\ \hline
10 & 25.0 & 67.6(+170.4\%)  & 25.6(\textbf{+2.4\%}) \\
20 & 43.2 & 105.8(+144.9\%) & 44.1(\textbf{+2.1\%}) \\
30 & 65.7 & 164.2(+149.9\%) & 67.4(\textbf{+2.6\%}) \\ \hline
\end{tabular}
\caption{Inference of a single forward pass measured in milliseconds, averaged over 100 times. 
}
\label{tenant_time}
\end{table}

For online applications, inference time is closely related to the actual user experience. Next we compare inference time of three algorithms versus different batch sizes. The results in Table \ref{tenant_time} show that AdapterDistillation has the same inference time as the Adapter method but it is significantly better than AdapterFusion. This is reasonable because an extra fusion layer in AdapterFusion takes some extra inference time. It is worth noting that the inference time of the Adapter/AdapterDistillation method is slightly larger (about 2.5$\%$) compared to full fine-tuning, which is caused by the serial insertion of the adapter module. Note that AdapterDistillation is independent of the structure of Adapter itself and can be hot-swapped into any Adapter-like method, such as Lora \citep{DBLP:conf/iclr/HuSWALWWC22}, to maintain the same inference time as full fine-tuning through parallel insertion.



In order to verify the improvement in model performance is due to the sharing of information from different tasks, we remove the current tenant from teacher adapters and train only using the existing tenants as teacher adapters. Figure \ref{fig:ablation_1} indicates that compared to adding the current tenant to the Teachers, the average accuracy is just slightly decreased by about 0.11$\%$, but still outperforms adapters by about 1.18$\%$. This indicates AdapterDistillation can effectively use multiple resources of extracted information.

\section{Conclusions and Future Work}
We proposed a novel and plug-and-play multi-adapter knowledge distillation algorithm called AdapterDistillation to implement the sharing of information between different tasks. Specifically, our proposed algorithm consisted of two stages of training. In the first stage of training, task-specific knowledge was extracted by training a student adapter using local data. Then in the second stage, knowledge from the existing teacher adapters was distillated into this student adapter through optimizing the distillation loss. Note that AdapterDistillation only employed the trained student adapter to implement inference,  which resulted in fast inference time and low resource consumption. Our proposed AdapterDistillation algorithm outperformed existing algorithms in terms of accuracy, resource consumption and inference time, meeting a practical scenario where numerous tenants accessing the platform in a streaming manner. In the future, plugging more advanced adapter structures into AdapterDistillation is an interesting direction to explore.




\bibliography{anthology,custom}
\bibliographystyle{acl_natbib}

\appendix


\section{Appendices}
\subsection{Detailed Experimental Setting}\label{Experimental_Setting}

In all experiments, we use Bert-base-Chinese\footnote{https://huggingface.co/bert-base-chinese} as the pre-training base model and set the classification threshold to be 0.5. 
All models are trained for 10 epochs with the same learning rate strategy as \citep{DBLP:conf/iclr/LoshchilovH19}. The distillation regularization parameter $\eta$ in (\ref{distill_explain}) is selected from [$e^{-2}$, $e^{-1}$, $e^{0}$,  $e^{1}$, $e^{2}$]. For  AdapterDistillation, we use the same parameter initialization strategy for all key, value, query matrices and the same hyper-parameters as AdapterFusion to ensure fair comparison. 
\subsection{Cold Start and Deployment}\label{sec:appendix}
Since some new tenants only have a knowledge base without any annotated data at the beginning, a universal pipeline for automatically building tenant datasets is proposed. The knowledge base contains many knowledge points, each of which corresponds to a standard question and multiple similar questions. Note that the collection of questions belonging to the same knowledge point has the same answer.

We automatically construct datasets through the following steps:

\noindent 1) Download knowledge base with the ID of the newly added tenant.\\
\noindent 2) Construct positive samples based on the labeled questions and similar questions in the knowledge base.\\
\noindent 3) Constructing negative samples using the BM25 algorithm \citep{DBLP:journals/ftir/RobertsonZ09}.\\

During the deployment of the service, we load adapter modules for all tenants on the pre-trained model. All requests from tenants on the platform will be directed to this model. During inference, the adapter module belonging to the corresponding tenant is activated based on their name, while those from other tenants are blocked.


\end{document}